\documentclass{article}
\usepackage{amsmath,graphicx,mlspconf}

\usepackage{amsmath,graphicx}
\usepackage[mathscr]{euscript}
\usepackage[Symbol]{upgreek}

\DeclareMathOperator{\tr}{Tr}
\newcommand {\vc} [1] {\boldsymbol{#1}}

\usepackage{algorithmic}
\usepackage{graphicx}
\usepackage{textcomp}
\usepackage{xcolor}
\usepackage{amsfonts,amsthm,bm, lipsum}
\def\BibTeX{{\rm B\kern-.05em{\sc i\kern-.025em b}\kern-.08em
    T\kern-.1667em\lower.7ex\hbox{E}\kern-.125emX}}
\title{Robust Learning via Ensemble Density Propagation in Deep Neural Networks}
\name{\hspace{-8mm} Giuseppina Carannante$^{\star }$, Dimah Dera$^{\star }$, Ghulam Rasool$^{\star },$ Nidhal C. Bouaynaya$^{\star }$, and Lyudmila Mihaylova$^{\dagger}$}
\address{\hspace{-8mm}$^{\star }$ Rowan University, Department of Electrical and Computer Engineering, 
    Glassboro, NJ\\
    \hspace{-8mm}$^{\dagger}$ University of Sheffield, Department of Automatic Control and Systems Engineering, United Kingdom\\
    \hspace{-8mm} $^{\star }$carannang1@rowan.edu, $^{\star }$derad6@rowan.edu, $^{\star }$rasool@rowan.edu,\\ $^{\star }$bouaynaya@rowan.edu, $^{\dagger}$l.s.mihaylova@sheffield.ac.uk}

\begin{document}

\maketitle

\begin{abstract}
Learning in uncertain, noisy, or adversarial environments is a challenging task for deep neural networks (DNNs). We propose a new theoretically grounded and efficient approach for robust learning that builds upon Bayesian estimation and Variational Inference. We formulate the problem of density propagation through layers of a DNN and solve it using an Ensemble Density Propagation (EnDP) scheme. The EnDP approach allows us to propagate moments of the variational probability distribution across the layers of a Bayesian DNN, enabling the estimation of the mean and covariance of the predictive distribution at the output of the model. Our experiments using MNIST and CIFAR-10 datasets show a significant improvement in the robustness of the trained models to random noise and adversarial attacks. 
\end{abstract}
\section{Introduction}\label{sec:intro}
Recently, machine learning models have shown significant success in various application areas, including computer vision and natural language processing \cite{krizhevsky2012imagenet,karras2019style}. However, these models may have limited suitability for mission-critical real-world applications due to the lack of information about the uncertainty (or equivalently confidence) in their predictions \cite{KendallGal2017UncertaintiesB}. Information about uncertainty and confidence can improve a model's robustness to random noise and adversarial attacks \cite{Ackerman,ker2017deep}. Many real-world applications, including various autonomous, military, or medical diagnosis and treatment systems, require the estimation of a model's confidence in its decisions \cite{Ackerman,ker2017deep}. Quantitative estimation of uncertainty in the model's prediction can be accomplished by exploiting well-established Bayesian methods.

In Bayesian settings, we start by defining a prior probability distribution over the unknown parameters, i.e., weights and biases of a DNN. Bayes' theorem allows us to infer the posterior distribution of these parameters after observing the training data \cite{Gal2016Bayesian, graves2011practical, Dera2019}. However, inferring the exact posterior distribution is mathematically intractable for most modern DNNs, as these models do not lend themselves to exact integration due to a large parameter space and multiple layers of nonlinearities \cite{bishop2006pattern}. One of the most common scalable density approximation approaches is Variational Inference (VI). The VI approximation method converts the intractable density inference into an optimization problem that is solved using standard algorithms, e.g., gradient descent \cite{bishop2006pattern, graves2011practical}. VI methods pose a simple family of distributions over the unknown parameters and then find (through optimization) a member of this family that is closest, in terms of Kullback-Leibler (KL) divergence, to the desired posterior distribution \cite{hinton1993keeping}. Over the past few years, VI has been used to estimate the posterior distribution for fully-connected neural networks, convolutional neural networks (CNNs), and recurrent neural networks \cite{blundell2015,shridhar2018,Gal2016Bayesian}. 

However, current Bayesian approaches based on VI do not propagate the variational distribution from one layer of the DNN to the next layer \cite{blundell2015}. Instead, a single set of parameters is sampled from the variational posterior and is used in the forward pass \cite{blundell2015}. Alternatively, the dropout is used at test time, mimicking a Bernulli distribution for the weights, to generate various samples, which, in turn, are used to calculate uncertainty in the output using the frequentist approach \cite{Gal2016Bayesian}.

Recently, Dera \emph{et al.} proposed a scalable and efficient approach, called extended VI (eVI), to propagate the first and second moments of the variational distribution through all layers of a CNN \cite{Dera2019, Dera2020radar}. Their method provided a mean vector and a covariance matrix at the output, corresponding to the network's prediction and uncertainty, respectively \cite{Dera2019}. The authors used first-order Taylor series approximation to compute the mean and covariance after propagating the variational distribution thorough the activation functions. However, the first-order Taylor series approximation may fail when the activation function is highly nonlinear, e.g., ELU, SELU, and Swish \cite{klambauer2017self,Clevert2016,ramachandran2017swish}. 

We build our Ensemble Density Propagation (EnDP) framework using the powerful statistical technique developed for density tracking in Ensemble Kalman Filters \cite{julier1997new}. We propagate random samples across the layers of DNNs and estimate the first two moments of the variational posterior after passing through each layer, including nonlinear activation functions. Our results show that propagating the variational posterior using EnDP results in increased robustness to Gaussian noise and adversarial attacks.

The rest of this paper is structured in the following way. In section \ref{sec:EnDP}, we describe the general VI framework and introduce our proposed Ensemble Density Propagation (EnDP) approach. EnDP results in the propagation of uncertainty information from the input, as well as networks parameters, to the network output. In section \ref{sec:results}, we present our results on a classification task using the MNIST and CIFAR-10 datasets and compare them to state-of-the-art VI approaches. In section \ref{sec:discussion}, we discuss our results and present the effect of the ensemble size (number of random samples N) on the performance of the proposed EnDP approach.

\section{Ensemble Density Propagation}
\label{sec:EnDP}
A framework for the propagation of the variational posterior density across layers of DNNs has been recently explored \cite{Dera2019}. In this paper, we introduce the Ensemble Density Propagation framework for tracking moments across layers of DNNs. We adopt the stochastic ensemble framework, drawing upon the ensemble Kalman filter and other Monte Carlo approaches \cite{fang2018nonlinear, van2001unscented}.  

We define a prior probability distribution $p(\mathrm{\Omega})$ over the set of weights $\mathrm{\Omega}$ of a DNN. After observing the training dataset $\cal{D}$, we update our belief and find the posterior distribution $p(\mathrm{\Omega}|{\cal{D}})$.
As the direct inference of $p(\mathrm{\Omega}|{\cal{D}})$ is intractable, we employ VI to approximate the true posterior with a parametrized distribution $q_{\vc{\theta}}(\mathrm{\Omega})$, also known as the variational posterior, with $\vc{\theta}$ representing the distribution parameters \cite{hinton1993keeping}. We assume $q_{\vc{\theta}}(\mathrm{\Omega})$ to be a Gaussian distribution. In VI, we minimize the KL-divergence between the true and the variational posterior distribution:
\begin{equation}\label{eq:1}
\mathbf{KL} (q_{\vc{\theta}}(\mathrm{\Omega})\big|\big|p(\Omega\big|\mathcal{D}))=
\int q_{\vc{\theta}}(\mathrm{\Omega}) \log\frac{  q_{\vc{\theta}}(\mathrm{\Omega})}{p(\mathrm{\Omega}) p( \cal{D} |\mathrm{\Omega})} d \mathrm{\Omega}.
\end{equation}

By rearranging the terms in (\ref{eq:1}), we obtain the following objective function:
\begin{equation}\label{eq:2}
\mathbf{\mathscr{L}}(\vc{\theta}) = -~\mathbb{E}_{q_{\vc{\theta}}(\mathbf{\Omega})}[\log(p(\mathrm{D}|\Omega)] + \mathbf{KL} (q_{\vc{\theta}}(\mathrm{\Omega})\big|\big|p(\Omega)),
\end{equation}
where $\mathbf{\mathscr{L}}(\vc{\theta}) $ is widely known as the {\it variational free energy} and is composed of two terms, the expected log-likelihood, which depends on the data, and the KL-divergence between the prior and variational posterior, which does not depend on the data and acts as a regularization penalty. 
For simplicity and without loss of generality, we present our EnDP framework for a single layer CNN with one max-pooling layer and a fully connected layer before the soft-max function. 
\paragraph*{Convolution Operation:}
In our framework, the convolutional kernels are assumed to be random variables endowed with a multivariate Gaussian distribution. We assume that the kernels within a convolutional layer are independent of each other. This assumption reduces the number of unknown parameters and also forces convolutional kernels to extract features that are uncorrelated with each other.

We consider the convolution operation as a matrix-vector multiplication. We express the output of the convolutional layer as $\bm{\mathbf{z}}^{(k_c)} = \mathbf{X}$ vec($\bm{\mathscr{W}}^{(k_c)}$), $k_c = 1, \cdots, K_c$, where $\mathbf{X}$ represents a matrix whose rows are the vectorized sub-tensors of the input image, $\bm{\mathscr{W}}^{(k_c)}$ is the $k_c^{\text{th}}$ convolutional kernel with vec$(\bm{\mathscr{W}}^{(k_c)})  \sim\mathcal{N}\left(\mathbf{m}^{(k_c)}, \mathbf{\Sigma}^{(k_c)}  \right)$, $K_c$ is the total number of kernels and (vec) is the vectorization operation. Thus, the output of the convolution between the $k_c^{\text{th}}$ kernel and the input image has a distribution $\bm{\mathbf{z}}^{(k_c)} \sim \mathcal{N}\left(\mathbf{X} \mathbf{m}^{(k_c)},\text{ } \mathbf{X} \mathbf{\Sigma}^{(k_c)} \mathbf{X}^T \right).$
\paragraph*{Nonlinear Activation Function:}
After the convolution, the resulting random variables $\bm{\mathbf{z}}^{(k_c)}$ will be propagated through an element-wise nonlinear activation function $\psi$. We perform stochastic sampling and draw $ N$ samples, $\bm{\mathbf{z}}^{(k_c)}_{i}$, where $i=1, 2, \cdots, N$. We pass each ensemble member $\bm{\mathbf{z}}^{(k_c)}_{i}$ through the activation function and obtain $\mathbf{g}_{i}^{(k_c)}=\psi[\bm{\mathbf{z}}^{(k_c)}_{i}]$. We find the sample mean and covariance of $\mathbf{g}^{(k_c)}$ using:
\begin{align}
    \bm{\mu}_{\mathbf{g}^{(k_c)}}&=\frac{1}{n}\sum_{i=1}^{N}\mathbf{g}_{i}^{(k_c)}, \\
    \bm{\Sigma}_{\mathbf{g}^{(k_c)}}&= \frac{1}{n-1}\sum_{i=1}^{N}\Big[\mathbf{g}_{i}^{(k_c)}-\bm{\mu}_{\mathbf{g}^{(k_c)}}\Big]\Big[\mathbf{g}_{i}^{(k_c)}-\bm{\mu}_{\mathbf{g}^{(k_c)}}\Big]^T \nonumber.
\vspace{-5mm}
\end{align}

\paragraph*{Max-Pooling Operation:}
The max-pooling operation selects the largest value in each patch of the given input. At the output of the max-pooling layer, we approximate the mean by $\bm{\mu}_{\mathbf{p}^{(k_c)}} = \text{pool}(\bm{\mu}_{\mathbf{g}^{(k_c)}})$. For the covariance matrix, we keep rows and columns of $\bm{\Sigma}_{\mathbf{g}^{(k_c)}}$ corresponding to the elements of the mean vector retained after the pooling operation, i.e.,  $\mathbf{\Sigma}_{\mathbf{p}^{(k_c)}} = \text{pool}(\mathbf{\Sigma}_{\mathbf{g}^{(k_c)}})$. 
If we denote by $d_{1} \times d_{1}$ the dimension of $\mathbf{g}^{(k_c)}$. Thus, $\bm{\mu}_{\mathbf{g}^{(k_c)}}$ has the same dimension as $\mathbf{g}^{(k_c)}$ and $\mathbf{\Sigma}_{\mathbf{g}^{(k_c)}}$ has a dimension $d_{1}^2 \times d_{1}^2$. At the output of max-pooling, the dimensions of $\bm{\mu}_{\mathbf{p}^{(k_c)}}$, and $\mathbf{\Sigma}_{\mathbf{p}^{(k_c)}}$ become $d_{2} \times d_{2}$ and $d_{2}^2 \times d_{2}^2$, respectively, where $d_{2}=(d_{1} - p)/s+1$, $p$ is the patch size of the pooling operation and $s$ is the stride.
\paragraph*{Fully-Connected (FC) Layer:}
The input to the FC layer $\mathbf{b}$ is obtained by vectorizing the output of the max-pooling layer. The mean and covariance of $\mathbf{b}$ are given by:
\begin{equation}
\hspace{-5 mm} \bm{\mu}_{\mathbf{b}}=\begin{bmatrix} \bm{\mu}_{\mathbf{p}^{(1)}} \\ \vdots \\ \bm{\mu}_{\mathbf{p}^{(K_c)}}\end{bmatrix}, \bm{\Sigma}_{\mathbf{b}}=\begin{bmatrix} \bm{\Sigma}_{\mathbf{p}^{(1)}}&\cdots&\mathbf{0} \\ \vdots& \ddots&\vdots \\ \mathbf{0}&\cdots&\bm{\Sigma}_{\mathbf{p}^{(K_c)}}\end{bmatrix}
\end{equation}
We denote the $h$\textsuperscript{th} weight vector of the FC layer by $\mathbf{w}_h \sim \mathcal{N}(\mathbf{m}_h, \bm{\Sigma}_h) $, for $h=1, \cdots, H$, where $H$ is the number of output neurons. By employing the derivations in \cite{Dera2019} for the product of random vectors, we can compute the output mean,  $\bm{\mu}_{\mathbf{f}}$,  and the output covariance,  $\bm{\Sigma}_{\mathbf{f}}$, of the FC-layer as:
\begin{align} \label{eq:fullyc}
\hspace{-1 mm} {\mu}_{f_{h}} &=  \mathbf{m}_h^T\bm{\mu}_\mathbf{b},\\
\bm{\Sigma}_{\mathbf{f}}  &= \begin{cases}
    \tr\big( \mathbf{\Sigma}_{h_i} \mathbf{\Sigma}_{\mathbf{b}}  \big) +\mathbf{m}_{h_i}^T  \mathbf{\Sigma}_{\mathbf{b}} \mathbf{m}_{h_j}+ \bm{\mu}_{\mathbf{b}}^T \mathbf{\Sigma}_{h_j} \bm{\mu}_{\mathbf{b}}, & i=j. \\
    \mathbf{m}_{h_i}^T  \mathbf{\Sigma}_{\mathbf{b}} \mathbf{m}_{h_j}, & i\neq j. \nonumber
  \end{cases}
\end{align}
where $h, h_i, h_j = 1, 2, \cdots, H$, and $i, j$ refer to any two weight vectors in the FC layer.
\paragraph*{Soft-max Function:}
For multi-class problems, the network output is given by the soft-max function, i.e., $\hat{\bm{y}}=\phi(\bm{f})$, where $\phi$ represents the softmax function and $\bm{f}$ is the output of the FC layer. We can approximate the mean $\bm{\mu}_{\mathbf{y}}$ and covariance $\bm{\Sigma}_{\mathbf{y}}$ using first-order Taylor series approximation:
\begin{equation}\label{eq:soft-max_output}
\bm{\mu}_{y} \approx \phi(\bm{\mu}_{f}), ~~ \text{and} ~~ \bm{\Sigma}_{\mathbf{y}} \approx \bm{J}_{\phi}\bm{\Sigma}_{\mathbf{f}}\bm{J}_{\phi}^T,
\end{equation}
\noindent{where $\bm{J}_{\phi}$ represents the Jacobian matrix of $\phi$ with respect to $\bm{f}$ evaluated at $\bm{\mu}_{f}$.}
The proposed EnDP framework can be easily extended to multi-layer CNNs and various architectures (such as recurrent neural networks) by following the same derivation presented above. 
\vspace{-2mm}
\section{EXPERIMENTS AND RESULTS}
\label{sec:results}
We evaluated the performance of the proposed EnDP method on a classification task, using two datasets, i.e., MNIST handwritten digits and CIFAR-10 \cite{Deng2012TheMD, Krizhevsky09}. We compared test accuracy of our model with the state-of-the-art in the literature, including a vanilla CNN, Bayes-by-Backprop (BBB), Bayes-CNN, Dropout-CNN, and eVI \cite{blundell2015, shridhar2018, Gal2016Bayesian, Dera2019}. We evaluated all models using test datasets of MNIST and CIFAR-10 under three conditions, i.e., noise-free, Gaussian noise, and adversarial attack. The targeted adversarial examples were generated using the fast gradient sign method (FGSM) \cite{Yanpei}.
\subsection{MNIST Dataset}
We used a CNN having one convolutional layer with 32 kernels of size $5 \times 5$, followed by the rectified linear unit (ReLU) activation, one max-pooling layer and one FC layer. We used $N = 1000 $ samples for the ensemble density propagation. We tested all models at two levels of Gaussian noise, i.e., $\sigma_{\text{noise}}^2=0.1, \text{and } 0.2$. The adversarial examples were generated to fool each model into predicting digit ``3''  with two attack levels, i.e., $\sigma_{\text{adversarial}}=0.1, \text{and } 0.2$.

\begin{figure*}[htb]
\begin{minipage}[b]{1.0\textwidth}
  \centering
  \centerline{\includegraphics[width=1\textwidth]{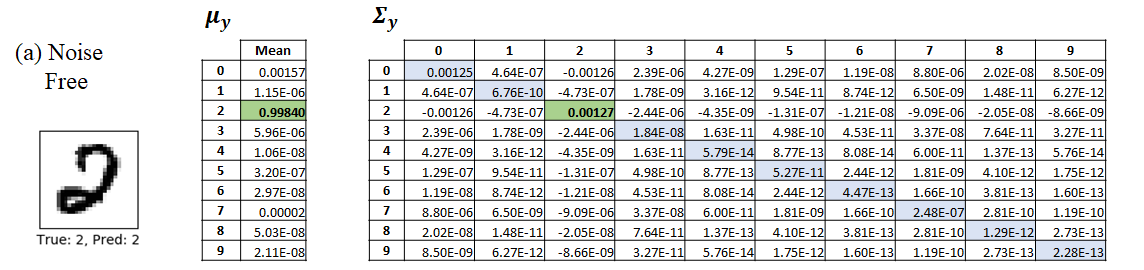}}
\end{minipage}
\begin{minipage}[b]{1.0\textwidth}
  \centering
  \centerline{\includegraphics[width=1\textwidth]{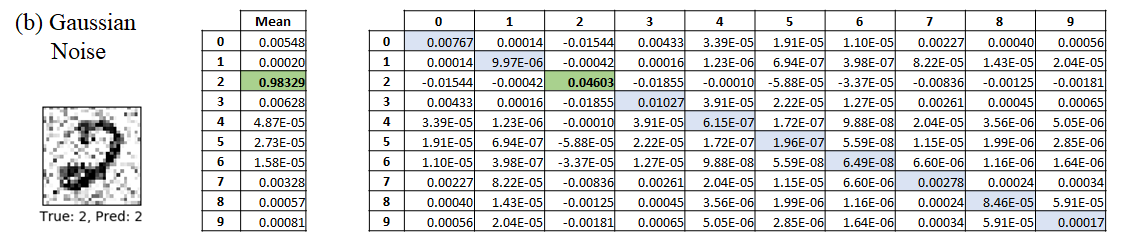}}
\end{minipage}
\begin{minipage}[b]{1.0\textwidth}
  \centering
  \centerline{\includegraphics[width=1\textwidth]{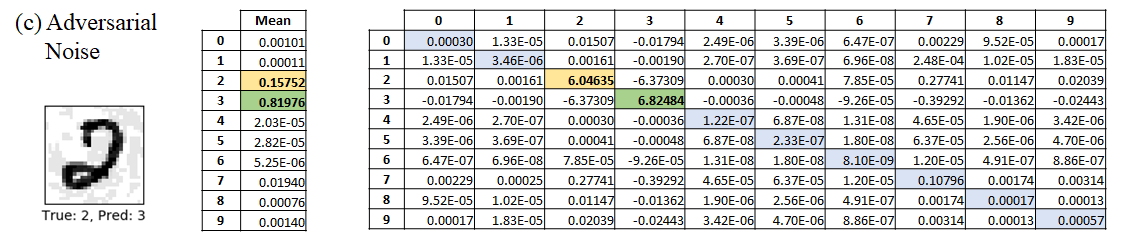}}
\end{minipage}
\caption{The output of the EnDP model, i.e., the mean vector $\vc{\mu}_{y}$ and covariance matrix $\bm{\Sigma_y}$ of the soft-max function, is presented for three test images. In sub-figures (b) and (c), test images were corrupted with Gaussian noise ($\sigma_{\text{noise}}^2=0.1$) and adversarial attack ($\sigma_{\text{adversarial}}=0.1$), respectively. The green color refers to the predicted output, while the yellow color represents the ground truth. When the yellow block is not shown, the network prediction and the ground-truth labels matched. In the covariance matrix, a large variance value indicates a low level of confidence or high uncertainty in the prediction.}
\label{fig:4}
\end{figure*}

In Table \ref{table:MNIST}, we present test accuracies of EnDP, eVI, BBB, and a vanilla CNN for the MNIST test set at various levels of Gaussian noise and adversarial attacks.
\begin{table}[b]
\caption{MNIST Test Accuracy}\label{table:MNIST}
\begin{center}
\begin{tabular}{c|c|c|c|c}
\hline
Noise/Attack level  & EnDP&eVI&BBB &CNN \\
\hline
\hline
No Noise  & 97\% & 96\% & 96\%  & 96\%\\
\hline
Gaussian Noise &  &  &  & \\
\hline
0.1  & 95\% & 94\% &  86\% & 79\%\\
\hline
0.2  & 86\% & 85\% &  76\% & 70\%\\
\hline
Adversarial Attack &  &  &  & \\
\hline
0.1  &95\% & 95\% &  91\% & 58\%\\
\hline
0.2 & 83\% & 81\% &  45\% & 14\%\\
\hline
\end{tabular}
\end{center}
\end{table}
In Fig. \ref{fig:4}, we present selected test results of EnDP for three different noise conditions, i.e., noise-free, Gaussian noise, and adversarial attack. We present test images with their ground-truth and predicted labels, and corresponding outputs of the soft-max function (the mean vectors $\vc{\mu}_{y}$ and covariance matrix $\bm{\Sigma_y}$ from Eq. \ref{eq:soft-max_output}). The diagonal elements of the covariance matrix, i.e., the variance elements, provide a meaningful and calibrated measure of the model's uncertainty or equivalently confidence associated with every prediction.

\begin{figure*}[ht]
\begin{minipage}{1\linewidth}
  \centering
 \centerline{\includegraphics[width=0.85\textwidth]{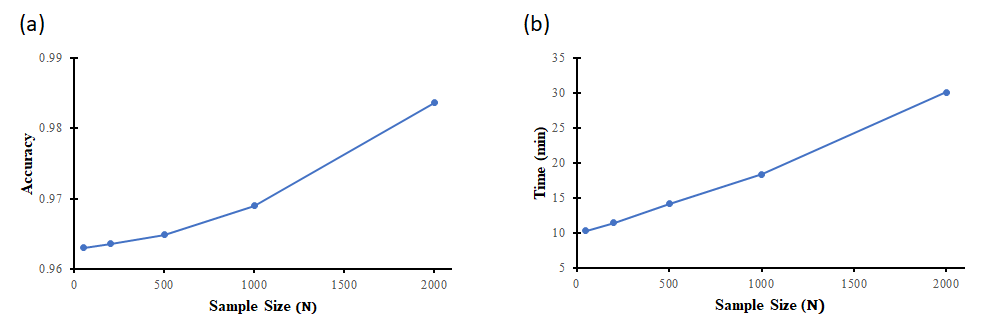}}
\end{minipage}
\caption{The effect of number of samples $N$ used for EnDP on the test accuracy and training time for MNIST dataset is presented. (a) Test accuracy increases as $N$ increases. (b) Training time (in minutes) for one epoch as $N$ is increased.}\label{fig:5}
\end{figure*}

In Fig. \ref{fig:5}, we present the test accuracy and training time of EnDP for various sample sizes $N$ used for ensemble density propagation.

\subsection{CIFAR-10 Dataset}
We used a CNN with three convolutional blocks and one FC layer. Each convolutional block included two consecutive convolutional layers, each followed by Exponential Linear Unit (ELU) activation function and one max-pooling layer at the end \cite{Clevert2016}. The convolutional kernels in all blocks were of size 3 $\times 3$. The number of convolutional kernels in the first, second and third block was set to $32$, $64$, and $128$, respectively. In total, our network included six convolutional layers, each followed by ELU activation.

For the ensemble density propagation, we used a different number of samples for each of the six ELU layers, i.e., $N_i =2d_i,$
where $i=1,2, \dots, 6$ represent ELU layers, and $d_i$ is the dimension of the feature map obtained after the $i$\textsuperscript{th} convolutional layer.
%
In Table \ref{table:CIFAR-10}, we report test accuracy of EnDP, eVI, Bayes-CNN and Dropout-CNN for the noise-free case and under adversarial and Gaussian noise conditions. The noise level was set to $5\%$ of the the highest conceivable value (HCV), where HCV $ = 3 ~ \sigma_{\text{noise}}$ \cite{duncan2000factors}. We generated the targeted adversarial examples to fool each network into predicting the label ``cat''.
\begin{table}[hb]
\caption{CIFAR-10 Test Accuracy}\label{table:CIFAR-10}
\begin{center}
\begin{tabular}{c|c|c|c|c}
\hline
Noise  & EnDP&eVI&Bayes- &Dropout- \\
Type & & &CNN & CNN\\
\hline
\hline
Zero & 86\% & 86\% & 85\%  & 86\%\\
\hline
Adversarial & 82\% & 80\% &  68\% & 52\%\\
\hline
Gaussian & 85\% & 82\% &  77\% & 75\%\\
\hline
\end{tabular}
\end{center}
\end{table}
\vspace{-2mm}
\section{DISCUSSION} \label{sec:discussion}
We proposed a new method for propagating variational posterior distribution through nonlinear activation functions in DNNs using the ensemble approach. We draw $N$ random samples, pass these samples thought the nonlinear activation functions, and calculate the mean and covariance of the transformed output. The propagation of the distribution through DNNs results in robust performance against Gaussian noise and adversarial attacks.

In the noise-free case, our approach, referred to as EnDP, performed better or equally on two benchmark datasets (MNIST and CIFAR-10) as compared to the state-of-the-art models, including eVI, BBB, Bayes-CNN, Dropout-CNN, and a vanilla CNN. Under noisy conditions and adversarial attacks, EnDP outperformed all models (except for the MNIST dataset at a low level of adversarial attack where EnDP and eVI produced 95\% test accuracy, Table \ref{table:MNIST}). We note that as the level of noise or severity of adversarial attack increased (Table \ref{table:MNIST}), the EnDP model maintained better performance. The gap between the accuracy of EnDP and other models increased. Similarly, in relatively complex network architecture (CIFAR-10 dataset, Table \ref{table:CIFAR-10}), EnDP performed robustly as compared to all other models in noise-free conditions as well as under noise attack. 

\subsection{Effect of Sample Size ($N$)}
We note that both the accuracy and training time increase with the increasing number of samples used for ensemble density propagation (Fig. \ref{fig:5}). This behavior agrees with the well-known trade-off between accuracy and computational cost. Our empirical results show that the number of samples required to achieve comparable accuracy depends upon the size of the feature map resulting from the preceding convolutional layer. We found that the number of samples approximately equal to twice the size of the feature map produced good results. For the case of MNIST, the output of the convolution operation $\bm{\mathbf{z}}$ is of size $d = 24 \times 24 = 576$. Therefore, we used $N = 1000$ for our experiments, which resulted in comparable accuracy at a reasonable computational cost. For CIFAR-10, we vary $N$ for each ELU layer depending upon the size of the output of the preceding convolutional layer ($N_i =2d_i$).
\subsection{Robustness to Noise and Adversarial Attacks}
We consider that the robustness of EnDP models to noise and adversarial attacks is attributable to the propagation of moments of the variational posterior through the network layers. The propagation of moments enables the model to use confidence (i.e., variance/covariance) information during the optimization process. In the moment propagation settings, the network learns ``robust'' parameters, including convolutional kernels and weights of the FC layer. The learned ``robust'' parameters result in a robust behavior, especially when the input is corrupted with noise or is adversarially attacked.

Both EnDP and eVI are based on variational posterior density propagation and show robustness in noisy and adversarial environments. However, the proposed EnDP method is superior to eVI, as evident in the experimental results, especially at a high level of noise and adversarial attacks. In our experiments, we used two activation functions, ReLU and ELU. However, the EnDP framework is readily extendable to all types of activation functions. Owing to the sampling and stochastic nature of our proposed EnDP technique, we consider that the performance of EnDP will be even better for highly nonlinear activation functions. In fact, we expect that for highly nonlinear activation functions (e.g., Gaussian Error Linear Unit, and Scaled exponential linear unit), the first-order approximation used in eVI might fail since higher-order terms are neglected in the linearisation; however, the proposed EnDP technique will perform robustly.
\subsection{Calibrated Uncertainty Information in the Model's Predictions}
The predictions of modern neural networks (i.e., the output of the soft-max function) are poorly calibrated and may provide misleading interpretation, especially when the predicted output is wrong \cite{guo2017calibration,nguyen2015deep}. A key feature of the proposed EnDP method is the availability of uncertainty information at the output through the covariance matrix. For example, consider the adversarial attack case in Fig. \ref{fig:4}(c). The EnDP model erroneously predicted digit ``3" instead of ``2"; however, the variance values (diagonal elements) corresponding to digits ``3" and ``2" were significantly larger as compared to all others. If we set the confidence proportional to the inverse of the variance, the mentioned example revels that the EnDP model was highly uncertain about its prediction and indicating low confidence in its output. The availability of a calibrated measure, i.e., the covariance matrix, can help establish the trustworthiness of machine learning models. Furthermore, the variance information can provide insights that can help interpret a model's correct and incorrect predictions.
\section{CONCLUSION} \label{sec:page}
We proposed a new approach for the approximation of variational posterior in DNNs. We were able to propagate the first two moments of the variational posterior through the layers of a multi-layer CNN using a stochastic ensemble technique. The proposed Ensemble Density Propagation (EnDP) framework can approximate any number of moments. The covariance matrix available at the output of an EnDP model captures its uncertainty in the predicted decisions. Our experimental results using the MNIST and CIFAR-10 datasets showed significantly increased robustness of the EnDP models to Gaussian noise and adversarial attacks. We consider that the propagation of moments through layers of the network results in robust learning and improved performance in noisy conditions.
\small{
\section{Acknowledgement}

This work was supported by the National Science Foundation Awards NSF ECCS-1903466, NSF CCF-1527822, NSF OAC-2008690. Giuseppina Carannante is supported by the US Department of Education through a Graduate Assistance in Areas of National Need (GAANN) Program Award Number P200A180055. We are also grateful to UK EPSRC support through EP/T013265/1 project NSF-EPSRC: ShiRAS. Towards Safe and Reliable Autonomy in Sensor Driven Systems.}
\bibliographystyle{IEEEbib}
 \bibliography{strings,refs}

\end{document}